\documentclass[journal]{IEEEtran}
\ifCLASSINFOpdf
\else
\fi

\usepackage{amsfonts}
\usepackage{amssymb}
\usepackage{amsmath}
\usepackage{bm}

\usepackage{algorithm}
\usepackage{algorithmic}

\usepackage{booktabs}
\usepackage{cite}

\usepackage{graphicx}
\usepackage{multirow}
\usepackage{hyperref}

\usepackage[flushleft]{threeparttable}

\begin{document}
%
\title{Tri-Branch Convolutional Neural Networks for Top-$k$ Focused Academic Performance Prediction}
%
%
%

\author{Chaoran~Cui,
        Jian~Zong,
        Yuling~Ma,
        Xinhua~Wang,
        Lei~Guo,
        Meng~Chen,
        and~Yilong~Yin
\thanks{
This work was supported by the National Natural Science Foundation of China under Grant 62077033 and Grant 61876098,
by the National Key R\&D Program of China under Grant 2018YFC0830100 and Grant 2018YFC0830102,
and by the Fostering Project of Dominant Discipline and Talent Team of Shandong Province Higher Education Institutions.}
\thanks{C. Cui is with the School of Computer Science and Technology, Shandong University of Finance and Economics, Jinan 250014, China (e-mail: crcui@sdufe.edu.cn).}
\thanks{J. Zong, M. Chen, and Y. Yin are with the School of Software, Shandong University, Jinan 250101, China (e-mail: superzj1111@163.com; mchen@sdu.edu.cn; ylyin@sdu.edu.cn).}
\thanks{Y. Ma is with the School of Computer Science and Technology, Shandong Jianzhu University, Jinan 250101, China (email: mayuling@mail.sdu.edu.cn).}
\thanks{X. Wang is with the School of Information Science and Engineering, Shandong Normal University, Jinan 250014, China (e-mail: wangxinhua@sdnu.edu.com).}
\thanks{L. Guo is with the School of Business, Shandong Normal University, Jinan 250014, China (e-mail: leiguo.cs@gmail.com).}
}

%
%

\markboth{Journal of \LaTeX\ Class Files,~Vol.~14, No.~8, August~2015}%
{Shell \MakeLowercase{\textit{et al.}}: Bare Demo of IEEEtran.cls for IEEE Journals}
%



\maketitle

\begin{abstract}
Academic performance prediction aims to leverage student-related information to predict their future academic outcomes, which is beneficial to numerous educational applications, such as personalized teaching and academic early warning.
In this paper, we address the problem by analyzing students' daily behavior trajectories, which can be comprehensively tracked with campus smartcard records.
Different from previous studies, we propose a novel Tri-Branch CNN architecture, which is equipped with row-wise, column-wise, and depth-wise convolution and attention operations, to capture the characteristics of persistence, regularity, and temporal distribution of student behavior in an end-to-end manner, respectively.
Also, we cast academic performance prediction as a top-$k$ ranking problem, and introduce a top-$k$ focused loss to ensure the accuracy of identifying academically at-risk students.
Extensive experiments were carried out on a large-scale real-world dataset, and we show that our approach substantially outperforms recently proposed methods for academic performance prediction.
For the sake of reproducibility, our codes have been released at \url{https://github.com/ZongJ1111/Academic-Performance-Prediction}.
\end{abstract}


%
\IEEEpeerreviewmaketitle

\section{Introduction}
%
%
%
%
\IEEEPARstart{S}tudent outcomes are the prioritized factors for higher educational institutions, and highly correlated with their reputation.
However, the risk of academic failure becomes an increasingly severe issue in education today.
According to the US National Center for Education Statistics~\cite{mcfarland2017undergraduate}, more than 30\% of college freshmen drop out after their first year of college.
Therefore, colleges constantly look for effective ways to enhance students' learning experiences.
Academic performance prediction aims to predict students' future academic performance based on student-related information~\cite{liu2019ekt}.
It not only facilitates personalized teaching, but also enables educators to provide timely interventions in learning, especially for those academically at-risk students~\cite{Macfadyen2010Mining}.

As one of the most popular topic in the field of educational data mining, there has been a large body of work on academic performance prediction during the past decades~\cite{Remero2010EDMreview}.
Earlier efforts were made by educational and psychological researchers, where they revealed essential factors affecting students' academic performance, including the big-five personality traits~\cite{Poropat2009A}, family environment\cite{kurdek1988relation}, and learning motivation~\cite{Fortier1995Academic}.
Nevertheless, these studies are based on students' questionnaires and self-reports, and probably suffer from the problems of small sample size and social desirability bias~\cite{yao2019predicting}.
Along another research line, many works~\cite{li2018different,huang2020learning} focused on e-learning platforms, such as MOOCs (Massive Open Online Courses), and predicted students' academic performance according to their online activities.
For example, it has been demonstrated that the final exam score is closely related to the number of assessments completed, video watching time, class discussion participation, etc., which, however, are hardly collected in off-line learning scenarios.

In educational psychology, it has been widely accepted that human behavior and learning ability are highly correlated~\cite{Wheaton2016School}.
From this point, with the growing advancement of campus IT applications, there emerge new opportunities to unobtrusively collect real-time records of students' campus behavior through smartphones~\cite{wang2014studentlife,wang2015smartgpa} and campus WiFi~\cite{zhou2016edum}, and understand students' behavioral patterns for academic performance prediction.
With the same motivation, students' daily studying and living activities have also been comprehensively sensed from campus smartcard records.
Previous studies~\cite{yao2017predicting,cao2018orderliness,yao2019predicting} resorted to manually extract high-level behavioral characteristics based on students' campus smartcard records, including orderliness, diligence, and sleep pattern, and used them as potential predictors for academic performance.
However, it is quite challenging to design effective handcrafted behavioral characteristics, and inevitably requires a considerable amount of domain expertise and engineering skills.

In this paper, we seek to reveal the students' behavior trajectories by mining campus smartcard records, and capture the characteristics inherent in trajectories~\cite{zhang2020social,liang2020learning} for academic performance prediction.
Inspired by the great success of deep learning in many fields, we propose an end-to-end prediction model based on Convolutional Neural Networks (CNNs).
Specifically, the behavior trajectory of each student is structured and compactly encoded as a third-order tensor, whose dimensions correspond to the date, time slot and campus location of individual activities, respectively.
Due to the capability of CNNs to hierarchically extract the spatial structural information, we carefully design a novel Tri-Branch CNN architecture, which is equipped with three types of convolutional kernels, i.e., the row-wise, column-wise, and depth-wise convolutions, to effectively capture the persistence, regularity, and temporal distribution characteristics of student behavior, respectively.
In addition, an attention module is separately appended to each convolution operation to adaptively weigh the impact of the behavioral characteristics at different dates, time slots, and campus locations on academic performance.

On the other hand, a primary goal of academic performance prediction is to enable the accurate and early identification of the students who are at risk of falling behind others, so that educators can subsequently offer guidance and helps in learning for these students.
However, existing works~\cite{yao2017predicting,cao2018orderliness,yao2019predicting} target at improving the prediction performance for the whole students, and pay less attention to the accuracy of detecting academically at-risk students.
To address this problem, we rank students in ascending order of their academic performance, and the top-$k$ ranked students exactly represent the at-risk ones with the worst performance.
As a result, academic performance prediction can be cast as a top-$k$ ranking problem~\cite{Niu2012}, which cares more about the accuracy of top-$k$ results.
Following the idea of cost-sensitive learning~\cite{khan2017cost}, we introduce a top-$k$ focused loss that assigns larger weights to training samples of students at higher positions, while smaller weights to those samples only involving students after the $k$-th position.
In this way, our algorithm gives priority to ensuring the correct identification of academically at-risk students.

To evaluate the capability of our approach, extensive experiments were conducted on a large-scale real-world dataset.
We presented detailed ablation studies on various configurations of our approach, and compared it against several recently proposed methods that also use campus smartcard records.
Besides, we visualized the trained models of our approach and provided insights into how it works for academic performance prediction.
Our codes have been released at \url{https://github.com/ZongJ1111/Academic-Performance-Prediction}.

To sum up, the main contributions of this paper are three-fold:
\begin{itemize}
  \item We propose a novel Tri-Branch CNN architecture to capture the characteristics of persistence, regularity, and temporal distribution from students' behavior trajectories.
      To the best of our knowledge, this is the first study to correlate students' campus behavioral patterns with academic performance in an end-to-end manner.
  \item We cast academic performance prediction as a top-$k$ ranking problem, in which the top-$k$ results correspond to academically at-risk students.
  By introducing a top-$k$ focused loss, we enable our algorithm to prioritize the accuracy of detecting the at-risk students.
  \item We demonstrate the effectiveness of our approach from different perspectives, and achieve the state-of-the-art results for academic performance prediction.
\end{itemize}

The remainder of the paper is structured as follows.
Section~\ref{sec:related} reviews the related work.
Section~\ref{sec:framework} details our framework for academic performance prediction.
Experimental results and analysis are reported in Section~\ref{sec:experiment}, followed by the conclusion in Section~\ref{sec:conclusion}.


\section{Related Works}\label{sec:related}
Academic performance prediction has emerged as a hot research topic due to its potential in numerous educational applications.
In the early stage, some physical and psychological factors were shown to have a significant correlation with students' academic performance, such as the health status~\cite{chang2002early,taras2005obesity}, intelligence quotient~\cite{deary2007intelligence}, and personality traits~\cite{Poropat2009A}.
However, it is generally difficult to change such factors so that students can be helped to improve their academic performance.
Moreover, these findings were derived from students' questionnaires and self-reports, which may be subject to a variety of inaccuracies and biases~\cite{yao2019predicting}.

In e-learning environments, academic performance prediction was realized according to students' online activity logs.
For example, Ren et al.~\cite{ren2016predicting} developed a personalized linear multi-regression model to predict the exam grades of a MOOC by considering the number of study sessions, quizzes, videos, and homework a student engages in.
He et al.~\cite{he2015identifying} built predictive models weekly for the students at risk of not completing a MOOC via two transfer learning based logistic regression algorithms.
Kim et al.~\cite{kim2018gritnet} regarded a student's learning activities as an event sequence, and input it to a bidirectional long short-term memory network to predict student graduation.
Li et al.~\cite{li2016dropout} used different types of students' learning behaviors to form multi-view features, and proposed a multi-view semi-supervised learning method for student dropout prediction.
Karimi et al.~\cite{karimi2020online} modeled the student course relations as a knowledge graph, and utilized a graph neural network to extract course and student embeddings to predict a student’s performance in a given course.

On the other hand, considerable research efforts were devoted to predicting students' performance as they work on online exercises.
Among them, the Item Response Theory (IRT)~\cite{embretson2013item} and Deterministic-Input Noisy-And-gate (DINA) model~\cite{de2011generalized} were widely applied to characterize students’ proficiency of different skills, and then the probability that a student correctly solves an exercise was modeled based on the association matrix between skills and exercises.
For both objective and subjective exercises, Liu et al.~\cite{liu2018fuzzy} proposed a fuzzy cognitive diagnosis framework under the fuzzy set theory and educational hypotheses, and simulated the generation of score on each exercise by considering slip and guess factors.
In addition, the problem was tackled using the matrix factorization algorithm~\cite{thai2012factorization}, which predicts the unknown scores of students given a student-exercise performance matrix with some historical scores.
As a step further, Liu et al.~\cite{liu2019ekt} presented an exercise-enhanced recurrent neural network by exploring both students' exercising records and the text content of corresponding exercises.

In off-line learning environments, the availability of detailed logs of students' learning activities poses a formidable barrier.
As a result, most previous methods on academic performance prediction relied on the demographic profiles~\cite{tamhane2014predicting,thammasiri2014critical}, past course grades~\cite{polyzou2019feature,ma2019pre}, in-class learning engagement~\cite{huang2013predicting,marbouti2016models}, homework practices~\cite{meier2015predicting}, etc., which can only be infrequently acquired from students.
Recently, students' behavioral data has received increasing attention, and been correlated with academic performance.
For example, Wang et al.~\cite{wang2014studentlife,wang2015smartgpa} exploited the smartphone sensing data to understand individual behavioral differences between high and low academic performers, and proposed a simple LASSO linear regression predictor for academic performance.
Zhou et al.~\cite{zhou2016edum} characterized students' punctuality for lectures using longitudinal WLAN data, and analyzed the relationship between punctuality and performance.
However, either smartphone sensing data or WLAN data needs to be specially collected from a small number of student volunteers.
Comparatively, Cao et al.~\cite{cao2018orderliness} and Yao et al.~\cite{yao2019predicting} tracked students' daily activities by mining campus smartcard records on a large scale, and extracted three high-level behavioral characteristics, i.e., orderliness, diligence, and sleep pattern, to predict the performance ranking of students.
Meanwhile, motivated by social influence theory, they found that behaviorally similar students have close academic performance, and incorporated student similarity to help the prediction of students without sufficient behavioral data~\cite{yao2017predicting}.
Lian et al.~\cite{lian2018jointly} studied academic performance prediction based on students' book-loan histories, and developed a supervised matrix factorization algorithm with multi-task learning for collaborative academic performance prediction and library book recommendation.

Despite the remarkable progress, all the above works manually designed students' behavioral characteristics in various ways, which are highly dependent on domain expertise and engineering skills.
By contrast, in this paper, we present a novel CNN-based deep learning model to automatically capture the characteristics inherent in students' behavior trajectories.
Additionally, similar to~\cite{cao2018orderliness,yao2019predicting}, we target at predicting the performance ranking of students, but give priority to ensuring the accuracy of top-$k$ results, which exactly refer to academically at-risk students in our study.


\section{Framework}\label{sec:framework}
In this section, we first formulate the problem of academic performance prediction by considering the students' campus behavior trajectories.
Then, we illustrate the Tri-Branch CNN architecture for academic performance prediction.
Finally, we introduce the top-$k$ focused loss to further optimize the accuracy of identifying academically at-risk students.

\begin{figure*}[t]
    \centering
    \includegraphics[width=0.85\textwidth]{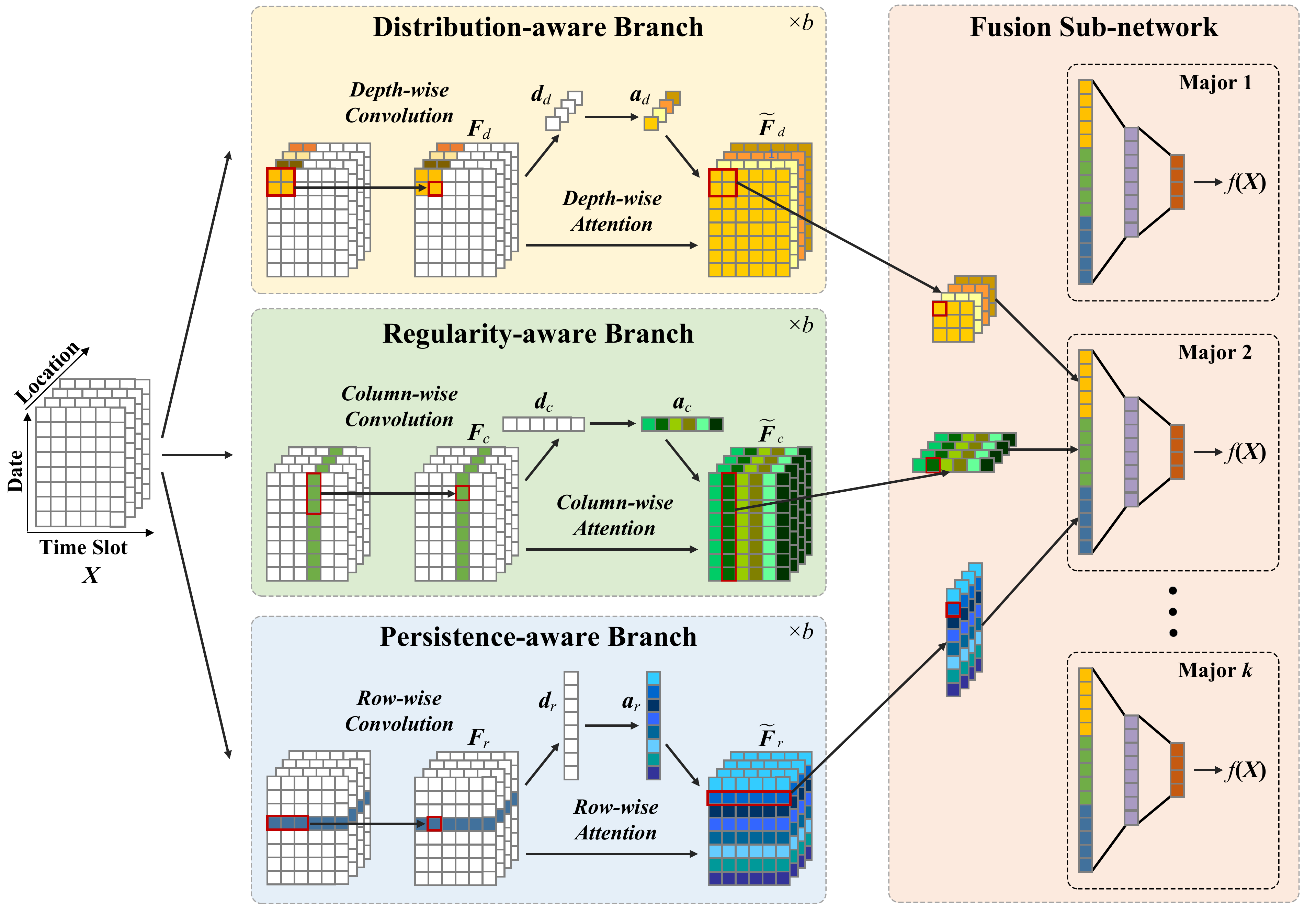}
    \caption{Architecture of the proposed Tri-Branch CNN for academic performance prediction.}
    \label{fig:architecture}
\end{figure*}

\subsection{Problem Formulation}

By mining the campus smartcard records, the students' behavior trajectories can be disclosed to some extent.
For each student, we collect the raw campus smartcard records of the student, and store a single record as a triple $(d_i, t_j, l_k)$, which indicates that the student appeared at the $k$-th campus location at the $j$-th time slot of the $i$-th day.
Furthermore, we represent the behavior trajectory of a student as a third-order tensor $\bm{X} \in \mathbb{R}^ {n_d \times n_t \times n_l}$, where $n_d$ denotes the number of days spanned by the records, $n_t$ denotes the number of time slots divided of a day, and $n_l$ denotes the number of campus locations involved in the records.
If the collected records of the student contain the triple $(d_i, t_j, l_k)$, we set $X(i,j,k)=1$, where $X(i,j,k)$ denotes the $(i,j,k)$-th entry of $\bm{X}$;
otherwise, $X(i,j,k)$ equals zero.

In this paper, we seek to capture the students' behavioral characteristics from their campus behavior trajectories, and correlate this knowledge with their academic performance.
Formally, the goal is to find a mapping function $f:{\mathbb{R}^ {n_d \times n_t \times n_l} \to \mathbb{R}}$, where given the trajectory representation of a student $\bm{X}$, $f(\bm{X})$ is a real value reflecting the performance level of the student.
However, it is rather difficult to directly estimate the absolute grade of academic performance of students;
instead, the relative comparisons between students can be judged more reliably~\cite{cao2018orderliness,yao2019predicting}.
Therefore, the training samples are provided in the form of pairwise comparisons of academic performance between students, i.e., $(\bm{X}_{u}, \bm{X}_{v}, y_{uv})$, where $\bm{X}_{u},\bm{X}_{v} \in \mathbb{R}^ {n_d \times n_t \times n_l}$ denote the representations of two students $u$ and $v$, and $y_{uv} \in \{ -1,+1 \} $ indicates their relative performance.
That is to say, $y_{uv} = +1$ indicates $u$ has better academic performance than $v$, and $y_{uv} = -1$ means the opposite.
The desired mapping function $f(\cdot)$ needs to be consistent with the pairwise comparisons between students, and can thus be learned with the hinge loss~\cite{peng2018discriminative}, i.e.,
\begin{equation}\label{eq:hinge_loss}
\sum_{(\bm{X}_{u}, \bm{X}_{v}, y_{uv}) \in \mathcal{D}}max(0,y_{uv}(f(\bm{X}_{v})-f(\bm{X}_{u}))+1)\ ,
\end{equation}
where $\mathcal{D}$ denotes the set of all training samples.

\subsection{Tri-Branch CNN}
We propose a novel CNN-based deep learning model, namely Tri-Branch CNN, as the instantiation of the mapping function $f(\cdot)$.
Fig.~\ref{fig:architecture} displays the architecture of the Tri-Branch CNN.
The network accepts the tensor representation $\bm{X}$ of each student, and feeds it into three parallel branches, i.e., persistence-aware branch, regularity-aware branch, and distribution-aware branch.
The three branches are equipped with row-wise, column-wise, and depth-wise convolution and attention operations, and play the role of capturing the persistence, regularity, and temporal distribution characteristics of student behavior, respectively.
Finally, the outputs of the three branches are merged via a fusion sub-network, in which the multi-task learning~\cite{shen2020deep} is introduced to separately yield the academic performance $f(\bm{X})$ of students from different majors.
In the following, we elaborate on each network branch and the fusion sub-network.

\subsubsection*{Persistence-aware Branch}
If a student stays in the same place for many hours of a day, it suggests that the student's behavior has the long persistence characteristic on that day.
To capture such information, we perform a row-wise convolution of kernel size $1 \times s$ in the persistence-aware branch.
Due to the ability of convolution operation to discover the local spatial structure of the data, the row-wise convolution models the changes of consecutive time slots on each day.
Moreover, the behavioral characteristics of students at different dates may have different effects on their academic performance.
For example, studying harder as you approach the final exam is more beneficial to achieve good performance.
Motivated by the observation, we append the row-wise attention mechanism after the convolution operation, which enables the network to adaptively determine the importance of different dates.
Denote by $\bm{F}_r \in \mathbb{R}^ {n_d \times n_t \times n_l}$ the feature maps outputted by each row-wise convolution in the persistence-aware branch.
The row context descriptor $\bm{d}_r \in \mathbb{R}^ {n_d \times 1}$ is generated by squeezing both time slot and location dimensions of $\bm{F}_r$:
\begin{equation}
{d_r}(i) = \frac{1}{{{n_t} \times {n_l}}}\sum\limits_{j = 1}^{{n_t}} {\sum\limits_{k = 1}^{{n_l}} {{F_r}(i,j,k)} },
\end{equation}
where $d_r(i)$ is the $i$-th element of $\bm{d}_r$, and ${F_r}(i,j,k)$ is the $(i,j,k)$-th entry of $\bm{F}_r$.
Then, $\bm{d}_r$ is used as the guidance to yield a soft attention map $\bm{a}_r \in \mathbb{R}^ {n_d \times 1}$ over the rows of $\bm{F}_r$.
This is realized by a simple two-layered perceptron with a bottleneck structure~\cite{hu2018squeeze}, i.e.,
\begin{equation}\label{eq:row_attention}
\bm{a}_r = MLP(\bm{d}_r),
\end{equation}
where the first layer performs dimensionality reduction with a reduction ratio $r$, while the second layer is to increase the dimension to $n_d$ again.
Finally, $\bm{a}_r$ is broadcast along the time slot and location dimensions,  and $\bm{F}_r$ can be recalibrated by the element-wise product with $\bm{a}_r$:
\begin{equation}
\widetilde{\bm{F}}_r = \bm{a}_r \otimes \bm{F}_r.
\end{equation}
In the persistence-aware branch, the block composed of row-wise convolution and attention operations is repeated $b$ times.
With the deepening of the branch, the local receptive field of the convolution kernel continues to expand.
As a result, the persistence characteristic over a long time interval of each day is effectively extracted.

\subsubsection*{Regularity-aware Branch}
If a student visits the same place at a fixed time every day, it means that the student's behavior has the strong regularity characteristic at that time.
Therefore, we use a column-wise convolution of kernel size $s \times 1$ in the regularity-aware branch, which captures the changes of consecutive days at the same time slot.
Also, the weights of different time slots are obtained by applying the column-wise attention mechanism.
Let $\bm{F}_c \in \mathbb{R}^ {n_d \times n_t \times n_l}$ be the feature maps generated by each column-wise convolution in the regularity-aware branch.
The column context descriptor $\bm{d}_c \in \mathbb{R}^ {n_t \times 1}$ can be acquired by shrinking $\bm{F}_c$ through the date and location dimensions:
\begin{equation}
{d_c}(j) = \frac{1}{{{n_d} \times {n_l}}}\sum\limits_{i = 1}^{{n_d}} {\sum\limits_{k = 1}^{{n_l}} {{F_c}(i,j,k)} }.
\end{equation}
Similar to Eq.~\eqref{eq:row_attention}, we use a two-layered perceptron to produce the soft attention map $\bm{a}_c$ over columns from $\bm{d}_c$, which sequentially conducts dimensionality reduction and increasing.
$\bm{a}_c$ needs to be broadcast along the date and location dimensions, so that the shapes of $\bm{F}_c$ and $\bm{a}_c$ are compatible for the subsequent element-wise multiplication, leading to the recalibrated feature maps $\widetilde{\bm{F}}_c$.
In the regularity-aware branch, we also repeat the operations of row-wise convolution combined with attention $b$ times, and thus measure the regularity characteristic over a long period of days for each time slot.

\subsubsection*{Distribution-aware Branch}
The presence of a student at different campus locations generally corresponds to his/her different behaviors, e.g., appearing in the library, supermarket and cafeteria implicitly expose the behavior of studying, shopping and eating, respectively.
Under the assumption of independence between the behaviors at different locations, we adopt the depth-wise convolution~\cite{howard2017mobilenets} in the distribution-aware branch, where a single convolution kernel of size $s \times s$ is performed over per channel of the input.
In this way, the temporal distribution pattern of each behavior over dates and time slots can be captured separately.
We denote by $\bm{F}_d \in \mathbb{R}^ {n_d \times n_t \times n_l}$ the feature maps computed from each depth-wise convolution.
The depth-wise attention mechanism is further imposed on $\bm{F}_d$ to weigh the impacts of the behaviors at different locations, in which the depth context descriptor $\bm{d}_d \in \mathbb{R}^ {n_l \times 1}$ is generated by:
\begin{equation}
{d_d}(k) = \frac{1}{{{n_d} \times {n_t}}}\sum\limits_{i = 1}^{{n_d}} {\sum\limits_{j = 1}^{{n_t}} {{F_d}(i,j,k)} }.
\end{equation}
Based on $\bm{d}_d$, the attention map $\bm{a}_d$ over depth is yield in an analogous way to that for $\bm{a}_r$ and $\bm{a}_c$.
$\bm{F}_d$ is then refined to generate $\widetilde{\bm{F}}_d$ by multiplying $\bm{a}_d$ in an element-wise manner.
Besides, the distribution-aware branch is the same as the other branches that contains $b$ blocks of depth-wise convolution and attention operations.

\subsubsection*{Fusion Sub-network}
At the end of persistence-aware, regularity-aware, and distribution-aware branches, we perform the max pooling on each row, column, and non-overlapped local region per feature map, leading to three outputs encoding the overall characteristics of behavioral persistence on each day, behavioral regularity at each time slot, as well as temporal distribution characteristic of the behavior at each location, respectively.
Then, they are flattened and concatenated, and passed to the fusion sub-network, which is designed as a simple three-layered fully-connected structure.
Through the multi-layer nonlinear transformations in the fusion sub-network, the features extracted from different network branches can be well balanced and merged for academic performance prediction.
Note that more complicated architectures may also be adopted for the fusion sub-network, but we leave them for future exploration.

Moreover, due to the inconsistent contents of academic performance tests across different majors, it is inappropriate to build the same model of academic performance prediction for all students.
Therefore, we regard the prediction for the students of each major as a single task, and introduce the multi-task learning~\cite{shen2020deep} to jointly solve different tasks in a unified framework.
As shown in Fig.~\ref{fig:architecture}, we follow the classic strategy of hard parameter sharing~\cite{luo2017heterogeneous}, that is, the three network branches are shared across different tasks, while the split takes place at the fusion sub-networks for task-specific predictions.
Through the multi-task learning, we are able not only to benefit from the relatedness between different tasks, but also to alleviate the problem of low number of students of some majors~\cite{ma2019multi}.

\subsection{Top-$k$ Focused Loss}
As aforementioned, a critical objective of academic performance prediction is to enable the accurate and early identification of students who are at risk of falling behind others, so that educators can provide in-time intervention and personalized guidance for the at-risk students~\cite{he2015identifying,marbouti2016models}.
Towards this end, we rank students in ascending order of their academic performance, and the top-$k$ ranked students exactly refer to the at-risk ones with the worst performance.
In our study, academic performance prediction is hence cast as a top-$k$ ranking problem~\cite{Niu2012}, which cares more about the accuracy of top-$k$ ranked students.

Specifically, we adopt the idea of cost-sensitive learning~\cite{khan2017cost}, and assign different weights to the training samples of pairwise comparisons between students.
Denote by $\bm{r}$ the true ranking of students in ascending of academic performance, whose element $r(u)$ indicates the ranking position of the student $u$.
Meanwhile, we define the gain that can be obtained when $u$ is ranked higher than others, which is reversely proportional to the position of the student in the true ranking, i.e.,
\begin{equation}\label{eq:gain}
  g(u) = n_s - r(u) + 1,
\end{equation}
where $n_s$ is the total number of students.
Following the popular metric of Discounted Cumulative Gain (DCG)~\cite{jarvelin2002cumulated} in information retrieval, given two students $u$ and $v$, the cumulative gain of $u$ and $v$ when they are arranged at the correct positions $r(u)$ and $r(v)$ can be computed by
\begin{equation}\label{eq:dcg}
{\rm{DC}}{{\rm{G}}_{uv}} = \frac{{g(u)}}{{{{\log }_2}\left(1+ {r(u)} \right)}} + \frac{{g(v)}}{{{{\log }_2}\left(1+ {r(v)} \right)}},
\end{equation}
in which the gain of each student is discounted by a logarithmic position factor.
Then, we exchange the positions of $u$ and $v$ in $\bm{r}$, and observe the change of DCG value, i.e.,
\begin{align}\label{eq:delta_dcg}
\Delta {\rm{DC}}{{\rm{G}}_{u \leftrightarrow v}} =&\ \left| {\frac{{g(u)}}{{{{\log }_2}\left( {1 + r(u)} \right)}} + \frac{{g(v)}}{{{{\log }_2}\left( {1 + r(v)} \right)}}} \right.\nonumber \\
&\ \left. { - \frac{{g(u)}}{{{{\log }_2}\left( {1 + r(v)} \right)}} - \frac{{g(v)}}{{{{\log }_2}\left( {1 + r(u)} \right)}}} \right|.
\end{align}
By substituting Eq.~\eqref{eq:gain} into Eq.~\eqref{eq:delta_dcg}, $\Delta {\rm{DC}}{{\rm{G}}_{u \leftrightarrow v}}$ can be rewritten in a concise form:
\begin{equation}\label{eq:delta_dcg_concise}
\left| {\left( {r(v) - r(u)} \right)\left( {\frac{1}{{{{\log }_2}\left( {1 + r(u)} \right)}} - \frac{1}{{{{\log }_2}\left( {1 + r(v)} \right)}}} \right)} \right|.
\end{equation}

Based on the above definitions, for the training sample $(\bm{X}_{u}, \bm{X}_{v}, y_{uv})$, the corresponding weight is empirically defined as
\begin{equation}\label{eq:weight}
{w_{uv}} = \left\{ \begin{aligned}
&\frac{{\Delta {\rm{DC}}{{\rm{G}}_{u \leftrightarrow v}}}}{{\Delta {\rm{DC}}{{\rm{G}}_{\max }}}}&&\mathrm{if}\ r(u) \le k\ \mathrm{or}\ r(v) \le k;\\
&\eta &&\mathrm{otherwise}.
\end{aligned} \right.
\end{equation}
Here, ${\Delta {\rm{DC}}{{\rm{G}}_{\max }}}$ is a normalization factor representing the maximum change of DCG value, which can be achieved when we exchange the first and last students in $\bm{r}$, i.e.,
\begin{equation}\label{eq:delta_dcg_max}
\Delta \mathrm{DC}{\mathrm{G}_{\max }} = ({n_s} - 1)\left( {1 - \frac{1}{{{{\log }_2}(1 + {n_s})}}} \right).
\end{equation}
Intuitively, if $u$ or $v$ belongs to the top-$k$ ranked students in $\bm{r}$, we take the normalized value of ${\Delta {\rm{DC}}{{\rm{G}}_{u \leftrightarrow v}}}$ as the sample weight $w_{uv}$.
Because the definition of ${\Delta {\rm{DC}}{{\rm{G}}_{u \leftrightarrow v}}}$ includes the position discount factor as shown in Eq.~\eqref{eq:delta_dcg_concise},
exchanging $u$ and $v$ at higher positions can lead to more change of DCG value, and a larger weight is assigned to the sample.
On the contrary, if both $u$ and $v$ appear after the $k$-th position in $\bm{r}$, there is no effect on the accuracy of top-$k$ results when exchanging their positions.
Therefore, we set $w_{uv}$ to be a small constant $\eta$.

Using the sample weights, a top-$k$ focused loss can be simply formulated by
\begin{equation}\label{eq:topk_focused_loss}
\sum_{(\bm{X}_{u}, \bm{X}_{v}, y_{uv}) \in \mathcal{D}}w_{uv}max(0,y_{uv}(f(\bm{X}_{v})-f(\bm{X}_{u}))+1).
\end{equation}
Since the training samples that involve students at higher positions have larger weights, the misclassification of these samples will result in larger losses and more backpropagated gradients;
instead, the training samples that only involve students after the $k$-th position have been assigned smaller weights, and they make less contribution during training.
As a result, through minimizing this new loss function, our approach gives priority to ensuring the accuracy of top-$k$ results in $\bm{r}$, that is to say, ensuring the correct identification of academically at-risk students.

\begin{table}[tbp]
\caption{Statistics of the dataset.}
\centering
\begin{tabular}{l c}
    \toprule
    \textbf{Statistics} & \\
    \midrule
    \# of students & 8,199 \\
    \# of majors & 19 \\
    Max. \# of students per major & 884 \\
    Min. \# of students per major & 85 \\

    \# of days\rule{0mm}{3mm} & 365 \\
    \# of time slots & 18 \\
    \# of locations & 12 \\

    Avg. \# of records of a student per day\rule{0mm}{3mm} & 5.57 \\
    \bottomrule
\end{tabular}
\label{tab:dataset}
\end{table}

\section{Experiments}\label{sec:experiment}
In this section, a series of experiments are conducted to evaluate our approach from different perspectives.
Through these experiments, we try to address the following Research Questions (RQ):
\begin{itemize}
\item \textbf{RQ1}: Does our approach benefit from the Tri-Branch CNN architecture?
\item \textbf{RQ2}: Does our approach uncover the effects of students' behavioral characteristics at different dates, time slots, and campus locations on academic performance?
\item \textbf{RQ3}: Does our approach yield better results by using the top-$k$ focused loss?
\item \textbf{RQ4}: Does our approach achieve the goal of improving the accuracy of academic performance prediction?
\end{itemize}

\subsection{Dataset}
The dataset used in this paper were collected from a public university in China.
It consists of two types of data, including the records of campus behavior and academic performance.
For privacy protection, all personally identifiable information is anonymized.
The main statistics of the dataset are summarized in Table~\ref{tab:dataset}.

\subsubsection*{Campus Behavior}
The dataset contains approximately 13.7 million campus smartcard records of 8,199 undergraduates from 19 majors covering an entire academic year, i.e., during 2014/09/01 to 2015/08/30.
The records reflect some consumption and entry-exit behaviors of students in campus, which totally take place at 12 different locations,
i.e., the laundry room, bathroom, teaching building, printing center, office building, library, cafeteria, school bus, supermarket, hospital, card center, and dormitory.
We only considered the students' behaviors occurring between 6am to 12pm of a day, and regarded each hour as a time slot.

\subsubsection*{Academic Performance}
The academic performance of each student is measured by Grade Point Average (GPA) over the academic year.
The absolute GPA scores were converted into the relative performance ranking of students within a major.
As mentioned before, we ranked students in ascending order of GPA scores.
In other words, students with poorer performance were arranged at higher ranking positions.

In our implementation, we randomly picked out 70\% of students from each major for training, 10\% for validation, and the remaining for testing.

\subsection{Evaluation Metric}
We measured the classification accuracy (Acc) of an algorithm on the pairwise comparisons of academic performance between students.
The Spearman Rank Correlation Coefficient~\cite{spearman1987proof} was also used as a ranking-based metric to quantify the correlation between the predicted and true performance ranking of students.
The Spearman's coefficient is defined as:
\begin{equation}
\rho  = 1 - \frac{{6\sum\nolimits_{u \in \mathcal{S}} {{{\left( {\hat{r}(u) - r(u)} \right)}^2}} }}{{\left| \mathcal{S} \right|\left( {{{\left| \mathcal{S} \right|}^2} - 1} \right)}},
\end{equation}
where $\hat{r}({u})$ and $r({u})$ are the predicted and true ranking positions of the student $u$, respectively, and $\mathcal{S}$ denotes the set of students in $u$'s major in the testing set.
Besides, we care about the correct identification of academically at-risk students, so the precision of top-$k$ ranked students (i.e., the $k$ students with the worst academic performance) was measured by:
\begin{equation}
p@k = \frac{{\left| {\mathcal{T}(\hat{\bm{r}},k) \cap \mathcal{T}(\bm{r},k)} \right|}}{k},
\end{equation}
where $\mathcal{T}(\hat{\bm{r}},k)$ and $\mathcal{T}(\bm{r},k)$ denote the sets of top-$k$ ranked students in the predicted ranking $\hat{\bm{r}}$ and true ranking $\bm{r}$, respectively.
$k$ was set to $10$ and $20$ in our case.
Note that the above metrics were separately measured for students in each major, and the average values over majors were reported to evaluate the overall performance.

\subsection{Implementation Details}
In this paper, we propose a Tri-Branch CNN architecture combined with the Top-$k$ Focused loss for academic performance prediction.
We refer to our approach as \textbf{TFTB-CNN}.
In the implementation of TFTB-CNN, the detailed experimental settings are described as follows:

In the Tri-Branch CNN architecture, we chose the kernel size $s=3$ for different convolution operators, and the reduction ratio $r=4$ for the attention module.
Each network branch contains 3 consecutive blocks of convolution and attention operations, i.e., $b=3$ in Fig.~\ref{fig:architecture}.
For the fusion sub-network, it consists of two hidden layers with 512 and 256 neurons, respectively, as well as an output layer with a single neuron.
For the top-$k$ focused loss, we set $k=10$ and $\eta=0.01$ in Eq.~\eqref{eq:weight}.

We implemented the network training and testing using the deep learning library Pytorch~\cite{paszke2019pytorch}.
The network was trained with the mini-batch Adam optimizer~\cite{kingma2014adam}.
We set the batch size to $16$.
The initial learning rate was $10^{-5}$ for all layers.
During training, we halved the learning rate every 20 epochs for a total of 50 epochs.

\begin{table}[tbp]
\caption{Performance comparison of the network architectures with different branches.}
\centering
\begin{threeparttable}
\begin{tabular}{l c c c c}
    \toprule
    Method & Acc & $\rho$ & $p@10$ & $p@20$  \\
    \midrule
    P-Branch\rule{0mm}{3mm} & 80.37 & 0.7915  & 36.32 & 57.63 \\ [2pt]
    R-Branch & 80.38 & 0.7936	& 39.47 & 59.47 \\ [2pt]
    D-Branch & 79.53 &	0.7572	& 40.15 & 59.21\\ [2pt]
    P-Branch + R-Branch	& 81.28 & 0.8073 & 37.89 & 61.32 \\ [2pt]
    P-Branch + D-Branch &	81.43	& 0.8113	& 41.05	& 61.58\\ [2pt]
    R-Branch + D-Branch	& 81.73	& 0.8127	& 42.11	& \textbf{62.89}	\\ [2pt]
    TB-CNN & \textbf{82.22}	& \textbf{0.8256} & \textbf{42.26} & 62.68 \\ [2pt]
    \bottomrule
\end{tabular}
\begin{tablenotes}
\footnotesize
\item The results in terms of accuracy (Acc), $p@10$, and $p@20$ are listed as percentage values. This convention is also adopted in the following tables and figures.
\end{tablenotes}
\end{threeparttable}
\label{tab:branch_ablation}
\end{table}

\subsection{Contribution of Network Branch}
In our approach, we design three network branches, i.e., persistence-aware branch (\textbf{P-Branch}), regularity-aware branch (\textbf{R-Branch}), and distribution-aware branch (\textbf{D-Branch}), to capture the persistence, regularity, and temporal distribution characteristics of student behavior, respectively.
To investigate the contribution of each branch, we compare the variants of the proposed Tri-Branch CNN (\textbf{TB-CNN}) architecture, which only integrate every single branch or their combinations thereof.
Note that a combination of any two branches is also realized via a three-layered fully-connected structure similar to the fusion sub-network as illustrated in Fig.~\ref{fig:architecture}.

Table~\ref{tab:branch_ablation} presents the performance comparison between the network architectures with different branches.
We can see that TB-CNN clearly outperforms its variants of fewer network branches in most metrics.
Meanwhile, combining any two branches frequently leads to higher performance over utilizing one of them alone.
The results suggest that the persistence, regularity, and temporal distribution characteristics of student behavior are complementary to each other for academic performance prediction, and the three branches of TB-CNN indeed improve the power of feature representation of the network.
Therefore, a positive answer to the research question \textbf{RQ1} can be formed.

\subsection{Benefit of Attention Mechanism}
In the Tri-Branch CNN architecture, we introduce the attention mechanism to adaptively weigh the impact of the behavioral characteristics at different dates, time slots, and campus locations on academic performance for each student.
Table~\ref{tab:attention_ablation} compares the Tri-Branch CNN against its variant without the attention modules, i.e., \textbf{TB-CNN versus TB-CNN w/o Att.}
Obviously, TB-CNN achieves better performance on all metrics, suggesting the benefit of applying the attention mechanism in the network.

\begin{table}[tbp]
\caption{The benefit of the attention mechanism.}
\centering
\begin{tabular}{l c c c c}
    \toprule
    Method & Acc & $\rho$ & $p@10$ & $p@20$  \\
    \midrule
    TB-CNN w/o Att.\rule{0mm}{3mm} & 81.91 & 0.8208  & 41.21 & 62.36 \\ [2pt]
    TB-CNN &  \textbf{82.22}	& \textbf{0.8256} & \textbf{42.26} & \textbf{62.68} \\ [2pt]
    \bottomrule
\end{tabular}
\label{tab:attention_ablation}
\end{table}

\begin{figure}[t]
    \centering
    \includegraphics[width=0.5\textwidth]{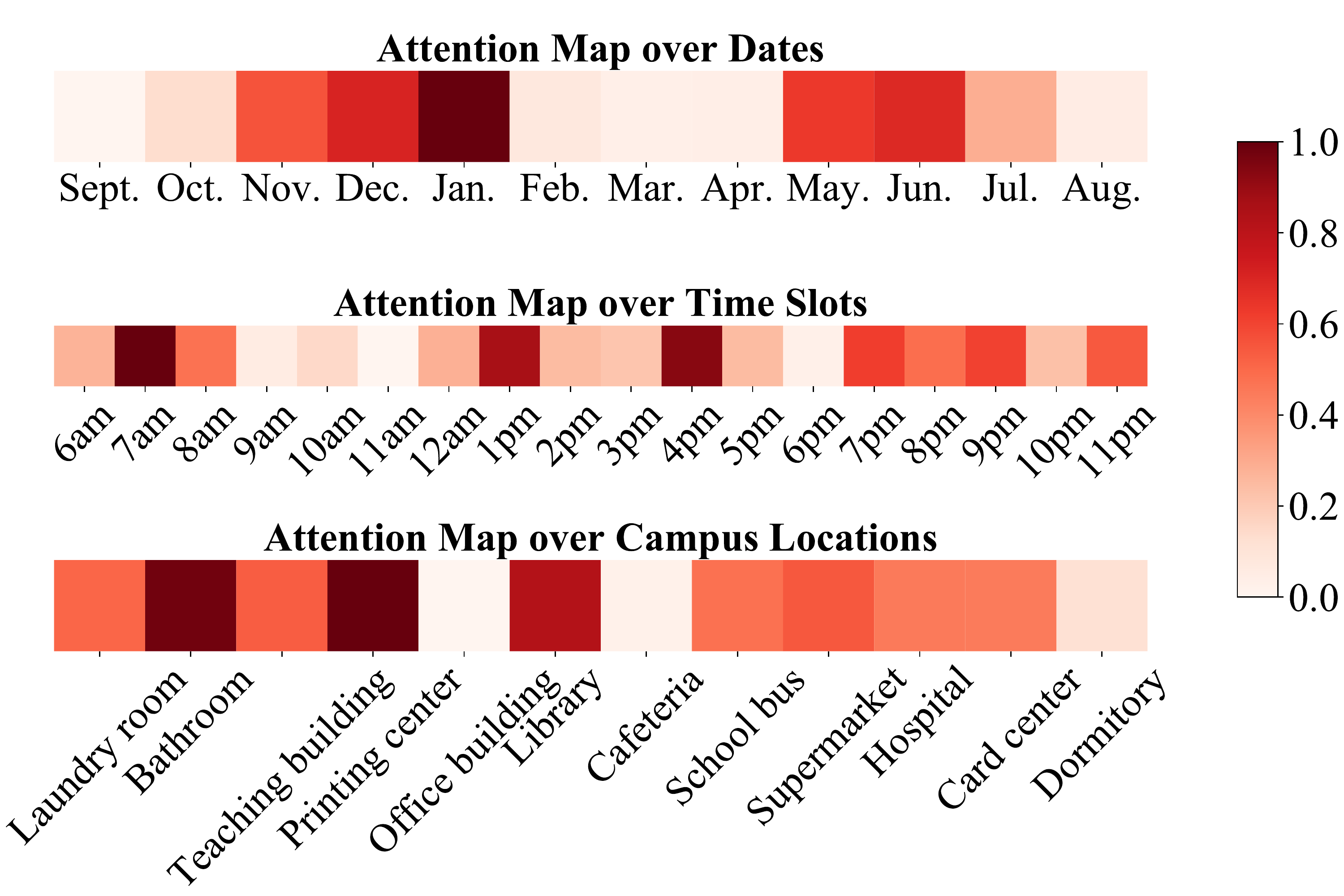}
    \caption{The attention map over dates, time slots, and campus locations, respectively.}
    \label{fig:attMap}
\end{figure}

To further gain insights into the role of the attention mechanism, in Fig.~\ref{fig:attMap}, we visualize the attention weights assigned to dates, time slots, and campus locations in the first block of network branches, which have been averaged over students.
Note that since the used campus smartcard records cover the whole 365 days in a year, we only show the mean of the attention weights of the days in a month for the convenience of visualization.
It can be seen that the network pays more attention to the periods from November to January as well as from May to July.
Both periods are close to the time of exams in universities, which are usually taken in January and July.
Therefore, the observations are consistent with our intuition that the behavioral patterns of the period approaching the exam have a greater impact on academic performance.
As for the time slots, it seems that how the students behave in non-class time, including the morning hours from 6am to 8am, the lunch break at 1pm, and the evening hours from 7pm to 9pm, can be regarded as a more important indicator for academic performance prediction.
The phenomenon is reasonable in the sense that students usually behave similarly with each other in class;
instead, their behavioral characteristics in spare time may provide more discriminative information.
At last, we can observe that the patterns of student presence in bathroom, printing center, and library affect more on academic performance.
A possible reason is that as revealed in~\cite{cao2018orderliness}, the records of taking showers can be used to quantify the orderliness of a student’s daily life.
On the other hand, students mostly go to printing center and library for study purposes.
Overall, the above results support the validity of introducing the attention mechanism from both quantitative and qualitative perspectives, and we can thereby give a positive answer to the research question \textbf{RQ2}.

\subsection{Effect of Top-$k$ Focused Loss}
As aforementioned, we seek to improve the accuracy of the top-$k$ results in the student ranking, that is, the $k$ students with the worst academic performance.
The top-$k$ focused loss is introduced as shown in Eq.~\eqref{eq:topk_focused_loss}, which assigns larger weights to the samples of students at higher positions and smaller weights to those samples involving only the students after the $k$-th position.
Table~\ref{tab:loss_ablation} displays the results of our approach with and without the top-$k$ focused loss, i.e., \textbf{TFTB-CNN versus TB-CNN}.
As can be seen, TFTB-CNN slightly outperforms TB-CNN in terms of Acc and $\rho$, but enjoys a distinct advantage in terms of $p@10$ and $p@20$.
For example, TFTB-CNN achieves about 7\% improvement on $p@10$ over TB-CNN.
This means that without compromising the prediction performance for the whole students, TFTB-CNN gives priority to ensuring the correct identification of academically at-risk students.
Therefore, we confirm the effectiveness of the top-$k$ focused loss, and provide a positive answer to the research question \textbf{RQ3}.
In addition, we notice that the superiority of TFTB-CNN is more obvious on $p@10$ than that on $p@20$.
The reason can be explained from two aspects:
1) We set $k=10$ in the top-$k$ focused loss, so the performance of TFTB-CNN in terms of $p@10$ is directly optimized during training;
and 2) As $k$ increases, the $p@k$ score generally increases for both methods.
In this situation, the performance gap between TFTB-CNN and TB-CNN may become narrower in terms of $p@20$.

\begin{table}[tbp]
\caption{The effect of the top-$k$ focused loss.}
\centering
\begin{tabular}{l c c c c}
    \toprule
    Method & Acc & $\rho$ & $p@10$ & $p@20$  \\
    \midrule
    TB-CNN\rule{0mm}{3mm} & 82.22 & 0.8256  & 42.26 & 62.68 \\ [2pt]
    TFTB-CNN & \textbf{82.77}	& \textbf{0.8282} & \textbf{49.28} & \textbf{64.14} \\ [2pt]
    \bottomrule
\end{tabular}
\label{tab:loss_ablation}
\end{table}

\begin{table*}[tbp]
\caption{Performance comparison of different methods when using the campus smartcard records covering different periods.}
\centering
\setlength{\tabcolsep}{1.2mm}{
\begin{tabular}{l c c c c c c c c c c c c c c c c c c c}
    \toprule
    \multirow{2}{*}{Method} & \multicolumn{4}{c}{30 days}& &\multicolumn{4}{c}{90 days} & &\multicolumn{4}{c}{180 days} & &\multicolumn{4}{c}{365 days} \\
    \cmidrule{2-5} \cmidrule{7-10} \cmidrule{12-15} \cmidrule{17-20}
    & Acc & $\rho$ & $p@10$ & $p@20$ & & Acc & $\rho$ & $p@10$ & $p@20$ & &  Acc & $\rho$ & $p@10$ & $p@20$ & &  Acc & $\rho$ & $p@10$ & $p@20$ \\
    \midrule
    Orderliness\rule{0mm}{3mm} & 57.08 & 0.2234  & 8.00 & 26.53 & & 56.36 & 0.2099 & 9.10 & 27.43 & & 56.73 & 0.1993 & 8.99 & 27.42 & & 57.45 & 0.2279 & 9.10 & 27.41 \\ [2pt]
    Diligence & 58.44 & 0.2762	& 7.90 & 26.66 & & 58.82 & 0.2098 & 7.79	& 26.47 && 60.45	& 0.3302	& 7.61	& 25.93	&& 59.39	& 0.3051	& 7.45 & 25.98 \\ [2pt]
    Sleep Pattern &	55.42 &	0.2071	& 6.85 & 25.42	&& 55.88	& 0.1828 &	7.26 &	25.57	&& 55.08	& 0.2144	& 7.49	& 25.94	&& 53.12	& 0.2369	& 7.37 &	25.94 \\ [2pt]
    RankNet	& 62.17 & 0.3626 & 8.10	& 26.92	&&  61.41	& 0.3439	& 9.00 & 27.66 &&	61.60	& 0.3548 &	9.31 &	28.13 &&	59.69	& 0.3186	& 7.71 & 25.72 \\ [2pt]
    LSTM &	73.44	& 0.6412	& 30.00	& 47.89 &&	75.93	& 0.6917	& 31.58 &	53.94	&& 77.66 &	0.7423	& 35.26	& 55.00 &&	79.66	& 0.7747	& 40.00	& 59.47 \\ [2pt]
    TCN	& 74.29	& 0.6419	& 32.63	& 50.53	&& 75.74	& 0.6736	& 38.42	& 53.42	&& 79.20 &	0.7505	& 39.47	& 58.16	&& 80.40 &	0.7849 &	41.05	& 60.00 \\ [2pt]
    TFTB-CNN & \textbf{79.43}	& \textbf{0.7627} & \textbf{38.01} & \textbf{60.36}	&& \textbf{81.04}	& \textbf{0.7899} &	\textbf{42.11}	& \textbf{62.21} &&	\textbf{81.42} &	\textbf{0.8084} &	\textbf{43.52}	& \textbf{62.84}	& & \textbf{82.77} &	\textbf{0.8282} &	\textbf{49.28} &	\textbf{64.14} \\ [2pt]
    \bottomrule
\end{tabular}}
\label{tab:performance_comparison}
\end{table*}

\subsection{Performance Comparison}
We compared our approach against the baseline methods presented in~\cite{cao2018orderliness,yao2019predicting}, where three high-level behavioral characteristics are manually designed based on students' campus smartcard records, i,e., orderliness, diligence, and sleep pattern.
Also, we experimented with the combination of the three behavioral characteristics:
\begin{itemize}
  \item \textbf{Orderliness}: The entropy of the temporal records of taking showers and having meals is computed as a proxy to quantify the orderliness of a student's daily life.
      The larger the entropy, the lower the orderliness, which is considered to be negatively correlated with a student's performance.
  \item \textbf{Diligence}: A student's cumulative occurrence of entering/exiting the library and fetching watering in the teaching building is used as a rough estimate of his/her diligence.
      The higher degree of diligence indicates the better performance.
  \item \textbf{Sleep Pattern}: The frequency of the first hour of smartcard records in each day is counted, and the timestamp of highest frequency is chosen to represent a student's wake-up time.
      The earlier a student wakes up, he/she is presumed to achieve the better performance.
  \item \textbf{RankNet}: The learning-to-rank model RankNet~\cite{burges2005learning} is trained to optimize the pairwise comparisons of academic performance between students by combining the above behavioral characteristics.
\end{itemize}

In addition, since a student's behavior trajectory is expressed as a sequence of visited campus locations, two classic deep sequence models were introduced to the comparison:
\begin{itemize}
  \item \textbf{LSTM}: The smartcard records during a single time slot is encoded as a binary $n_l$-dimensional vector, where the $i$-th element indicates the appearance at the $i$-th campus location.
      The encodings of consecutive time slots for a student are sequentially fed into a LSTM network~\cite{gao2017identifying}, which is trained in the same way as RankNet.
  \item \textbf{TCN}: Similar to LSTM, the Temporal Convolutional Network (TCN) architecture~\cite{bai2018empirical} is also used for the sequential modeling of student behavior along time slots.
\end{itemize}
For the LSTM model, we chose the number of hidden layers and the number of cells per layer to be 2 and 64, respectively.
The TCN model was implemented with 2 residual blocks, each of which contains 3 dilated convolutional layers with the kernel size of 2 and dilation rates of 1, 2, and 8, respectively.

\begin{figure*}[tbp]
    \centering
    \includegraphics[width=0.95\textwidth]{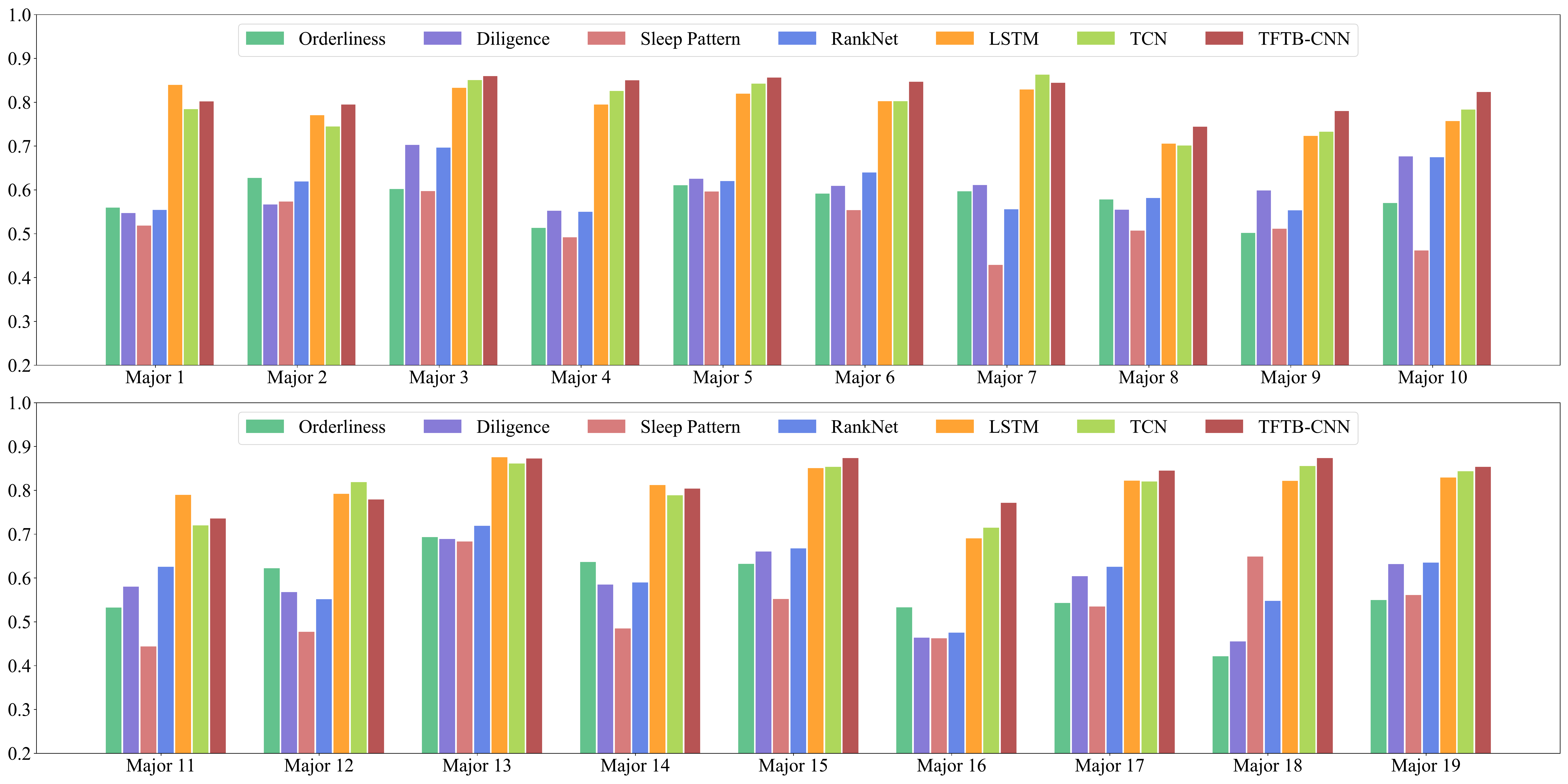}
    \caption{Performance comparison between the methods across students of different majors on Acc when using all records of 365 days.}
    \label{fig:histogram}
\end{figure*}

We studied how these methods perform when using the smartcard records covering different periods, i.e., the first 30 days, 90 days, 180 days, and all 365 days.
The settings simulate the scenarios that we predict the annual academic performance depending on the clues of students' campus behavior trajectory only from the first month, quarter, half year and the full year, respectively.
Table~\ref{tab:performance_comparison} lists the performance of the proposed TFTB-CNN compared to the baseline methods.
It can be seen that TFTB-CNN is substantially better than its contenders in all conditions.
For example, it exceeds the second place by nearly 2.3\% with respect to Acc in both conditions of 180-day and 365-day records.
The results suggest that TFTB-CNN is more effective than the other competitors for academic performance prediction, and provide the evidence that the research question \textbf{RQ4} can be positively answered.
Also, the following important observations can be made from Table~\ref{tab:performance_comparison}:
\begin{itemize}
  \item In the comparison, Orderliness, Diligence, Sleep Pattern, and RankNet, are all based on the manually designed behavioral characteristics.
      They considerably fall behind the deep learning based methods, including LSTM, TCN, and TFTB-CNN.
      This highlights the merit of the end-to-end learning techniques for academic performance prediction.
  \item All methods experience a degradation in performance with the decrease of the time period covered by the used records.
        However, TFTB-CNN exhibits a greater relative advantage with this change.
        For example, the improvement of $\rho$ arises from at least 0.04 in the condition of 365-day records to 0.12 in that of 30-day records.
        The results suggest that TFTB-CNN can somewhat relieve the problem of data insufficiency.
        It is worth noting that even with only the limited records from the first 30 days, TFTB-CNN still maintains about 80\% classification accuracy and a value of $\rho$ over 0.75.
        We thus believe that TFTB-CNN has a higher capability of predicting students' future academic outcomes in the earlier stage.
  \item In terms of $p@10$ and $p@20$, TFTB-CNN is significantly ahead of the other competitors.
        This observation again verifies the necessity of introducing the top-$k$ focused loss for detecting academically at-risk students, and supports the positive answer to the research question \textbf{RQ3}.
\end{itemize}
In Fig.~\ref{fig:histogram}, we further examined the performance of methods across students of different majors on Acc when using all records of 365 days.
As expected, TFTB-CNN consistently exceeds the others in most cases, exhibiting its high robustness regarding the change of students' majors.

\subsection{Visualization of Student Embeddings}
\begin{figure*}[tbp]
    \centering
    \includegraphics[width=0.81\textwidth]{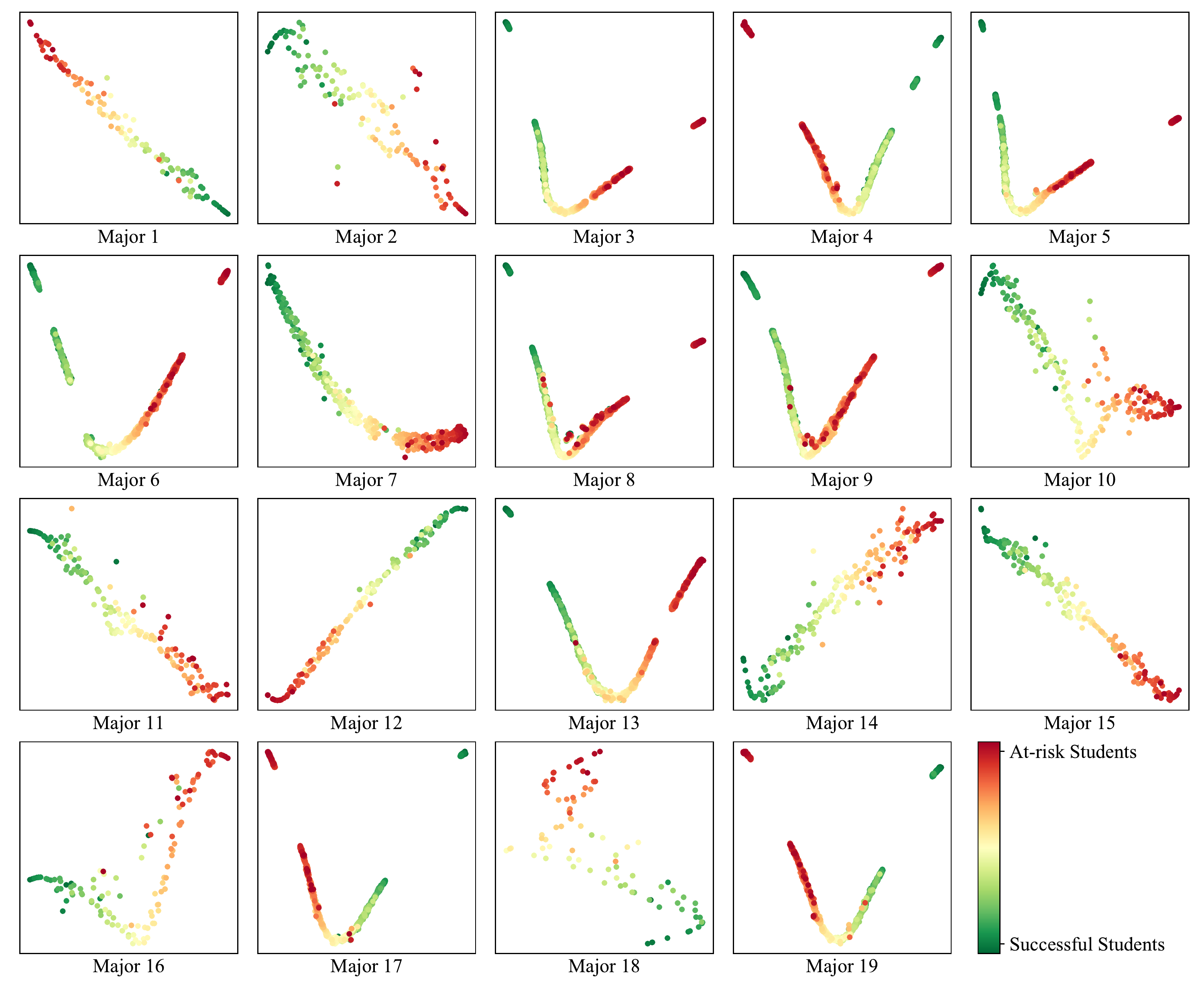}
    \caption{Visualization of the learned embeddings of students from different majors.}
    \label{fig:embedding}
\end{figure*}

In academic performance prediction, the key issue is to learn the appropriate feature representation to discriminate between academically successful and at-risk students.
To validate whether the two groups are visually distinguished in the learned feature space, we took the embedding vectors of students from the last fully-connected layer of the network, and projected them into a two-dimensional space using the t-SNE algorithm~\cite{maaten2008visualizing}.
Fig.~\ref{fig:embedding} shows the projection results of student embeddings from different majors.
Obviously, the embeddings of academically successful and at-risk students can be well separated for all majors.
Besides, the embeddings are approximately distributed over a manifold structure, where the students with similar academic performance are still close to each other.
From the above findings, we demonstrate that our approach has the powerful ability of representation learning for academic performance prediction.

\section{Conclusions}\label{sec:conclusion}
Inspired by the observation that human behavior and learning ability are highly correlated, we show the possibility of discovering students' behavior trajectories from campus smartcard records and leveraging the cues to predict their academic performance.
In our framework, we present an end-to-end Tri-Branch CNN model that performs row-wise, column-wise, depth-wise convolution and attention operations to capture the characteristics of persistence, regularity, and temporal distribution of student behavior, respectively.
Academic performance prediction is considered as a top-$k$ ranking problem, for which we introduce a top-$k$ focused loss to prioritize the correct identification of academically at-risk students.
In the experiments, we verify the effectiveness of the key components of our approach, including the network branches, attention modules, and loss function, through detailed ablation studies.
We also demonstrate that our approach consistently and substantially outperforms benchmark methods under different conditions for academic performance prediction.

Due to the difficulty of data acquisition, some important factors related to students' academic performance are not explored in this study, e.g., historical test scores.
In future research, we plan to collect more relevant data and combine them with students' daily behavioral records.
Besides, as pointed out in prior works~\cite{yao2017predicting,feng2019understanding}, students may have similar academic performance if they behave in a similar fashion.
Therefore, we will investigate how to measure the students' behavioral similarity and leverage the underlying relationships between students to extend our current model.


%




\ifCLASSOPTIONcaptionsoff
  \newpage
\fi

\begin{IEEEbiography}[{\includegraphics[width=1in,height=1.25in,clip]{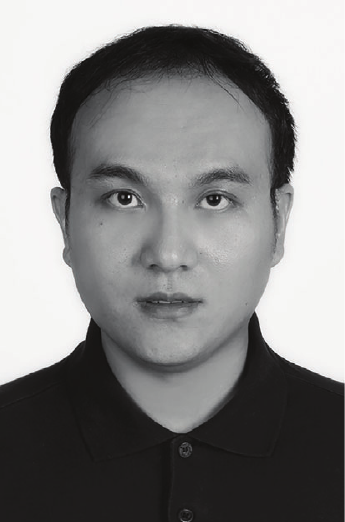}}]{Chaoran Cui}
received his Ph.D. degree in computer science from Shandong University in 2015. Prior to that, he received his B.E. degree in software engineering from Shandong University in 2010. During 2015-2016, he was a research fellow at Singapore Management University. He is now a professor with School of Computer Science and Technology, Shandong University of Finance and Economics. His research interests include information retrieval, recommender systems, multimedia, and machine learning.
\end{IEEEbiography}
\begin{IEEEbiography}[{\includegraphics[width=1in,height=1.25in,clip]{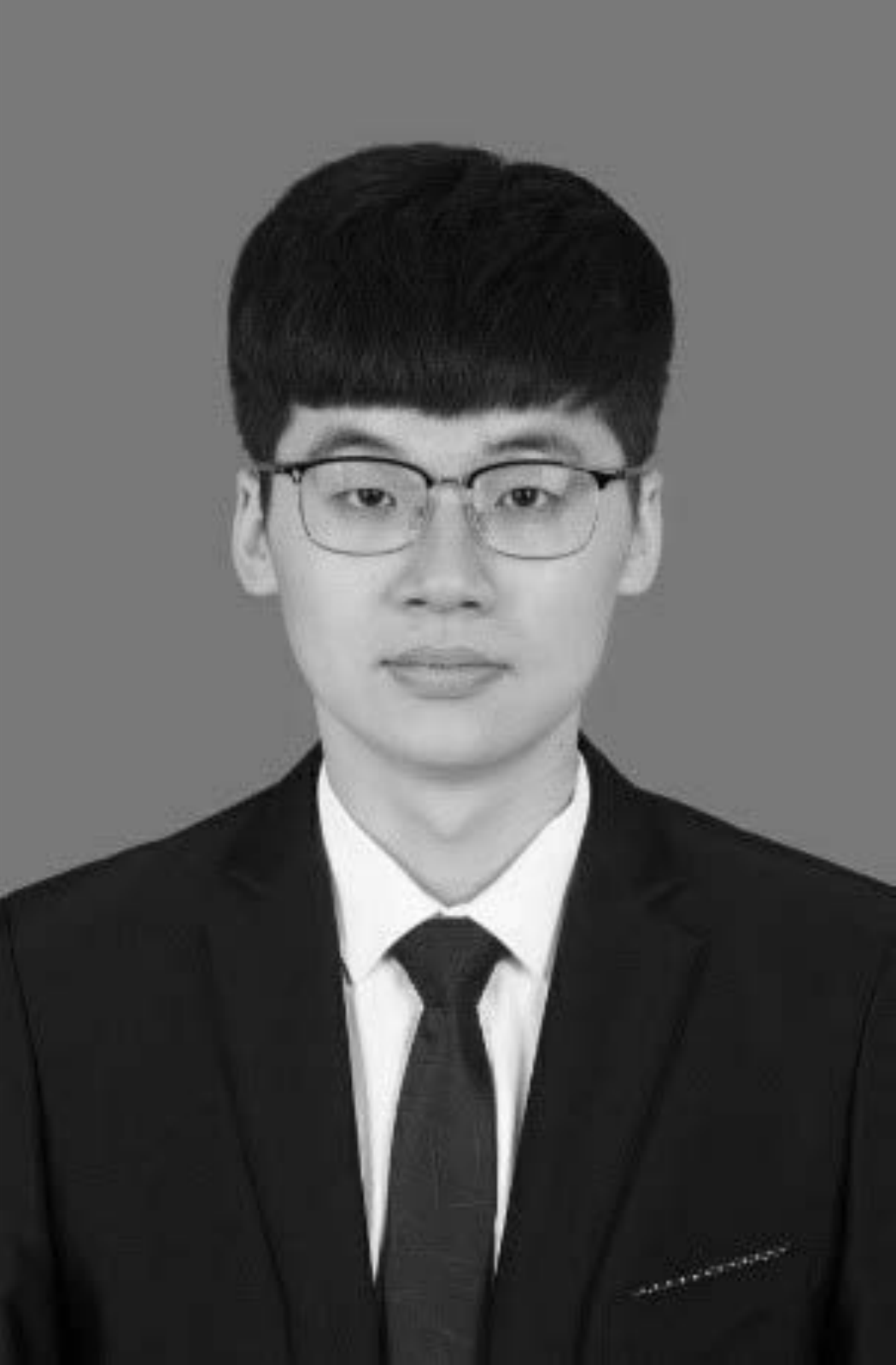}}]{Jian Zong}
received his B.E. degree in network engineering from Henan University, Kaifeng, China, in 2018. He is currently pursuing the master degree in software engineering at Shandong University, Jinan, China. His research interests include educational data mining and machine learning.
\end{IEEEbiography}
\begin{IEEEbiography}[{\includegraphics[width=1in,height=1.25in,clip]{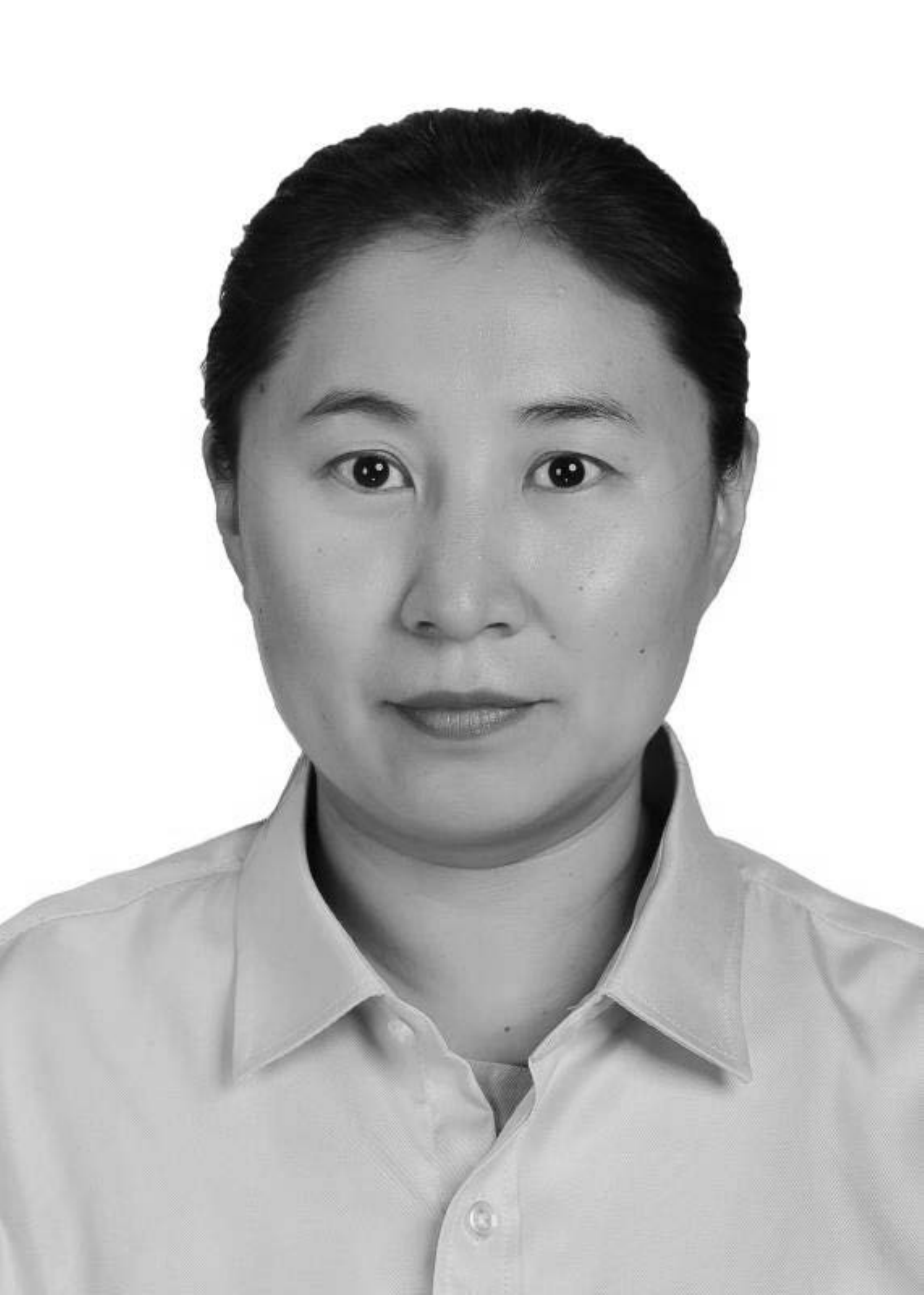}}]{Yuling Ma}
received her Master degree and Ph.D. degree in computer science and technology from Shandong University, China, in 2008 and 2020, respectively. She is currently a Lecturer with the School of Computer Science and Technology, Shandong Jianzhu University. Her research interests are educational data mining and machine learning.
\end{IEEEbiography}
\begin{IEEEbiography}[{\includegraphics[width=1in,height=1.25in,clip]{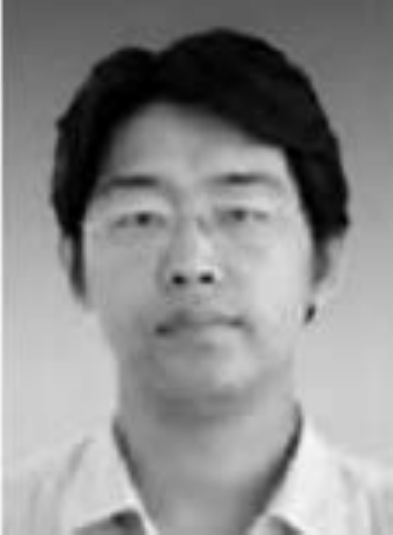}}]{Xinhua Wang}
received the M.S. degree in applied mathematics from the Dalian University of Technology, China, in 2001, and the Ph.D. degree in management science and engineering from Shandong Normal University, China, in 2008. He was a Visiting Scholar with Shandong University, from 2002 to 2003, and a Senior Visiting Scholar with Peking University, from 2008 to 2009. He is currently a Professor and a Master Supervisor with the School of Information Science and Engineering, Shandong Normal University. His research interests include distributed networks and recommender systems.
\end{IEEEbiography}
\begin{IEEEbiography}[{\includegraphics[width=1in,height=1.25in,clip]{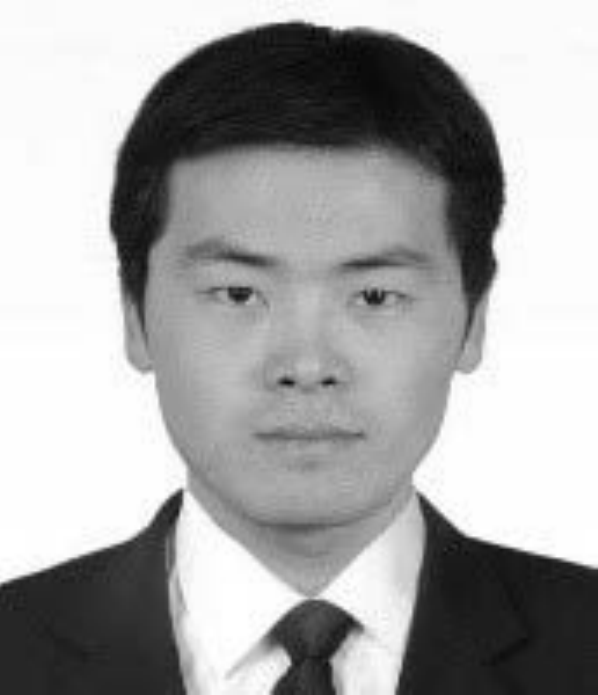}}]{Lei Guo}
received the Ph.D. degree in computer architecture from Shandong University, China, in 2015. He was a Visiting Scholar with The University of Queensland (UQ), Australia, from 2018 to 2019. He is currently an Associate Professor and a Master Supervisor with Shandong Normal University, China. His research grants include the National Natural Science Foundation of China and the China Postdoctoral Science Foundation. His research interests include information retrieval, social networks, and recommender systems.
\end{IEEEbiography}
\begin{IEEEbiography}[{\includegraphics[width=1in,height=1.25in,clip]{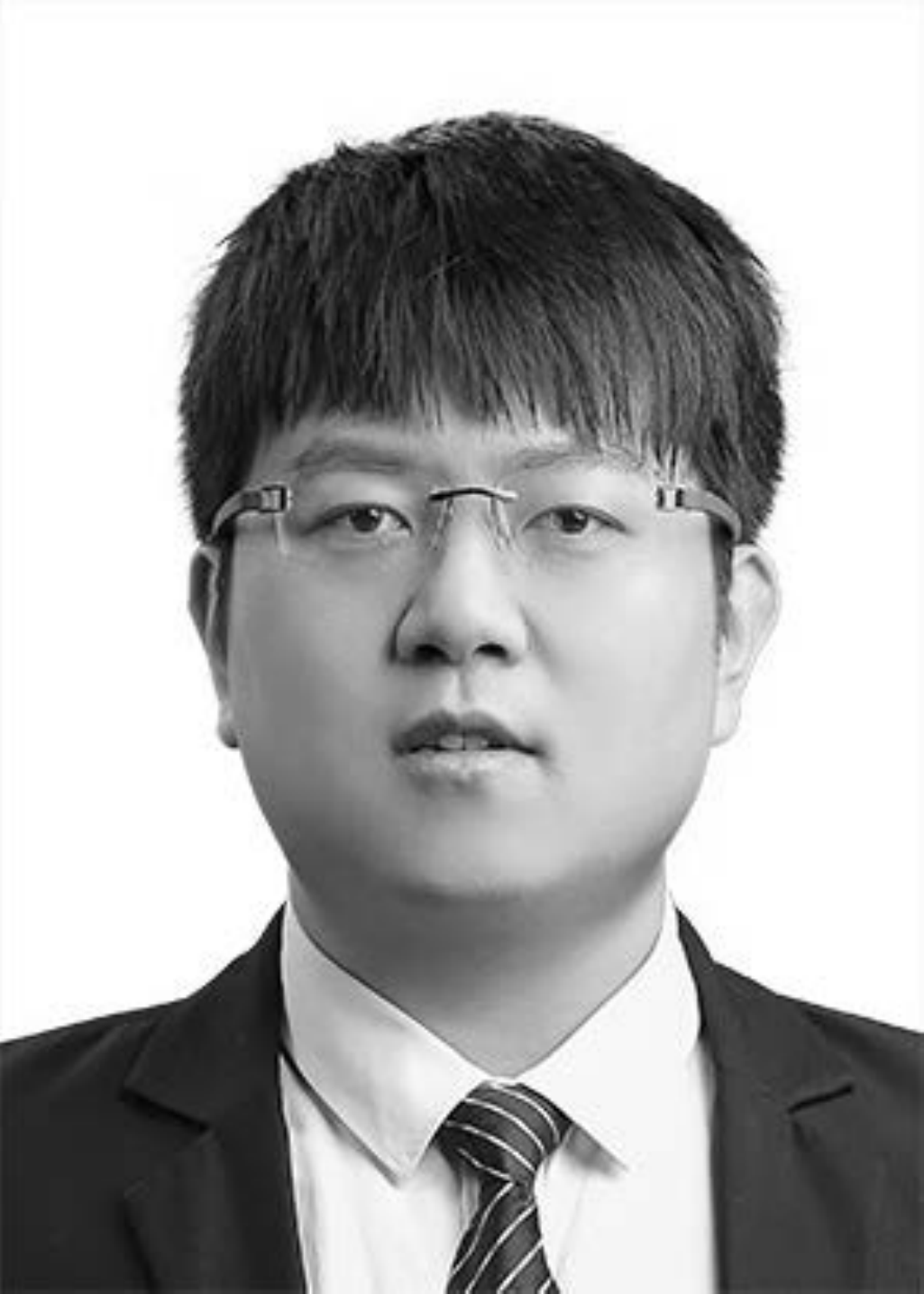}}]{Meng Chen}
received his Ph.D. degree in computer science and technology in 2016 from Shandong University, China. He worked as a Postdoctoral fellow from 2016 to 2018 in the School of Information Technology, York University, Canada. He is currently an assistant professor in the School of Software, Shandong University, China. His research interest is in the area of trajectory data mining and traffic management.
\end{IEEEbiography}
\begin{IEEEbiography}[{\includegraphics[width=1in,height=1.25in,clip]{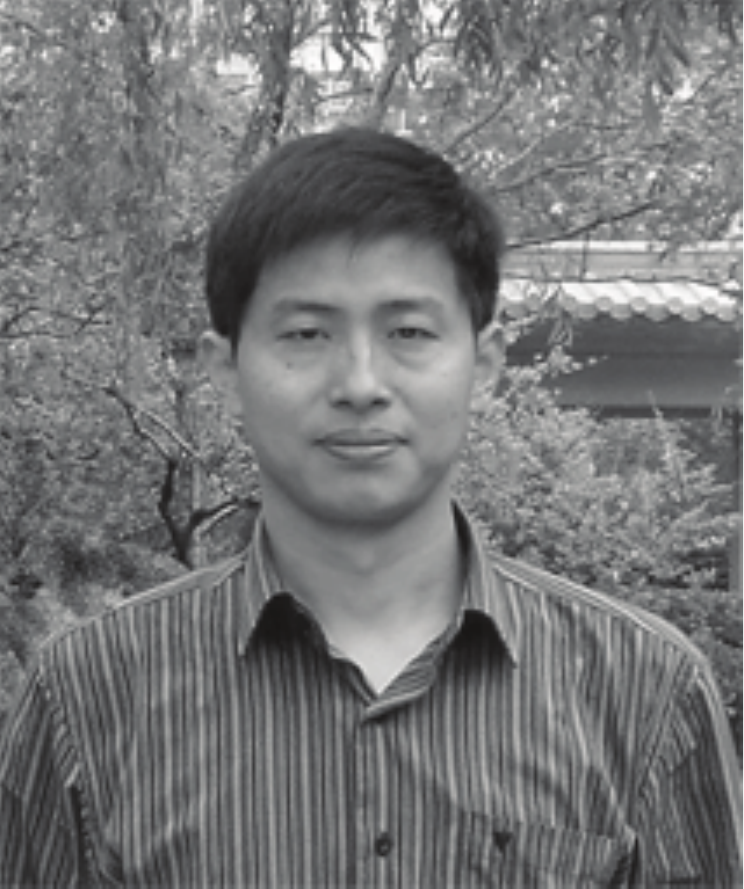}}]{Yilong Yin}
received the Ph.D. degree from Jilin University, Changchun, China, in 2000. He is currently the Director of the Machine Learning and Applications Group and a Professor with Shandong University, Jinan, China. From 2000 to 2002, he was a Postdoctoral Fellow with the Department of Electronic Science and Engineering, Nanjing University, Nanjing, China. His research interests include machine learning, data mining, computational medicine, and biometrics.
\end{IEEEbiography}



\vfill


\end{document}